\documentclass[numbers]{article}

\usepackage[main, final]{neurips_2025}
\usepackage{algorithm}
\usepackage{algorithmic}
\usepackage{listings}
\usepackage{xcolor}
\usepackage{amsmath}
\usepackage{makecell}
\usepackage{graphicx}
\usepackage{subcaption}
\usepackage{caption}
\lstdefinestyle{mystyle}{
  language=Python,
  numbers=left,
  numberstyle=\tiny,
  basicstyle=\ttfamily\small,
  keywordstyle=\bfseries,
  commentstyle=\color{green!60!black},
  columns=flexible,
  breaklines=true,
  showstringspaces=false,
  frame=single,
  captionpos=b
}

\usepackage[utf8]{inputenc} 
\usepackage[T1]{fontenc}    
\usepackage{hyperref}       
\usepackage{url}            
\usepackage{booktabs}       
\usepackage{amsfonts}       
\usepackage{nicefrac}       
\usepackage{microtype}      
\usepackage{xcolor}         

\title{Contrastive Learning for Multi-Label ECG Classification with Jaccard Score–Based Sigmoid Loss}

\author{%
  Junichiro Takahashi \quad 
  Masataka Sato \quad 
  Satoshi Kodeta \quad 
  Norihiko Takeda \\
  Department of Cardiovascular Medicine \\
  The University of Tokyo Hospital\\
  Tokyo, Japan\\
  \texttt{kodera@tke.att.ne.jp} \\
}

\begin{document}

\maketitle

\begin{abstract}
Recent advances in large language models (LLMs) have enabled the development of multimodal medical AI. While models such as MedGemini achieve high accuracy on VQA tasks like USMLE-MM, their performance on ECG-based tasks remains limited, and some models, such as MedGemma, do not support ECG data at all. Interpreting ECGs is inherently challenging, and diagnostic accuracy can vary depending on the interpreter’s experience. Although echocardiography provides rich diagnostic information, it requires specialized equipment and personnel, limiting its availability. 

In this study, we focus on constructing a robust ECG encoder for multimodal pretraining using real-world hospital data. We employ SigLIP, a CLIP-based model with a sigmoid-based loss function enabling multi-label prediction, and introduce a modified loss function tailored to the multi-label nature of ECG data. Experiments demonstrate that incorporating medical knowledge in the language model and applying the modified loss significantly improve multi-label ECG classification. To further enhance performance, we increase the embedding dimensionality and apply random cropping to mitigate data drift.

Finally, per-label analysis reveals which ECG findings are easier or harder to predict. Our study provides a foundational framework for developing medical models that utilize ECG data.
\end{abstract}

\section{Introduction}
In recent years, alongside the emergence of large language models (LLMs), multimodal medical AI has been developed. Recently, models such as MedGemini~\cite{saab2024capabilitiesgeminimodelsmedicine} and MedGemma~\cite{sellergren2025medgemmatechnicalreport} have been introduced, marking the appearance of multimodal models in the medical domain. However, while MedGemini achieves high accuracy on VQA tasks such as USMLE-MM, reaching 93.5\%, its performance on ECG-QA, which involves electrocardiogram data (ECG), is considerably lower at 57.7\%. In addition, MedGemma does not support an ECG at all. This discrepancy can be attributed to the inherently challenging nature of ECGs for model training. 

In real-world clinical settings, interpreting ECGs is one of the more challenging tasks, and it is well known that diagnostic accuracy can vary significantly depending on the interpreter’s professional background and level of experience~\cite{kashou2023ecg}. Although transthoracic echocardiography is recommended for the diagnosis of cardiovascular diseases due to its rich informational content~\cite{heidenreich20222022}, it requires specialized technicians, and many facilities lack sufficient infrastructure to perform the examination~\cite{wood2014left}. In this context, the development of a multimodal model capable of handling electrocardiogram data and estimating echocardiographic findings from ECGs could provide substantial support in clinical settings. However, to date, no such clinically useful multimodal model exists.

To build a high-quality multimodal model, it is essential to design ECG encoders suitable for the modality. In this study, we focus on the construction of a convincing encoder for ECGs. Previous studies have reported attempts to apply 
CLIP~\cite{radford2021learningtransferablevisualmodels} has been used as a pretraining method for ECGs~\cite{cai2025supremesupervisedpretrainingframework, li2023frozenlanguagemodelhelps, yu2024ecgsemanticintegratoresi}, but these approaches have several limitations. First, many of these studies rely on publicly available datasets such as PTB-XL~\cite{strodthoff2023ptb} rather than real-world clinical data. While previous studies have made predictions using simplified PTB-XL labels, actual clinical findings are more finely categorized, and other predictive factors present in a single ECG can be overlooked. Therefore, a more detailed classification is necessary.
Second, while real-world cardiovascular diseases often involve multiple abnormalities simultaneously, representing a multi-label problem, existing studies applying CLIP have been limited to single-class prediction. Therefore, CLIP-based contrastive learning for ECGs may not capture the inherent multi-label characteristics of ECG data.

In this study, we employed real-world hospital data and conducted pretraining based on SigLIP~\cite{zhai2023sigmoid}, assessing its performance in multi-label prediction tasks. SigLIP is a model that replaces the CrossEntropyLoss of CLIP with a sigmoid-based loss function, thereby enabling multi-label inference for each prediction. We also demonstrate that improving the loss function is necessary to enhance multi-label classification performance when training ECG data using SigLIP. Moreover, we addressed the clinically significant task of estimating echocardiographic findings from ECGs, investigating the potential of ECGs as a surrogate for echocardiography.

Overall, our study introduces two principal contributions. First, it leverages authentic clinical data for multimodal pretraining, enhancing the clinical validity of the model. Second, it adopts a sigmoid-based loss function to facilitate multi-label prediction, thereby enabling clinically meaningful inferences from ECGs that were not achievable with previous CLIP-based approaches.

\section{Methods}
\subsection{Model architecture}
In this study, we trained an ECG encoder using SigLIP and evaluated its performance in multi-label classification. The predicted findings are presented in the Appendix~\ref{appendix:labels}. We adopted a 1D ResNet-18 as an ECG Encoder as previous studies~\cite{cai2025supremesupervisedpretrainingframework, li2023frozenlanguagemodelhelps} have reported superior performance compared with Vision Transformer (ViT) architectures. As the language model, we employed Qwen3-8B~\cite{qwen3technicalreport}, which was selected based on preliminary evaluation indicating a favorable balance between model size and domain-specific knowledge regarding the target labels. For the ablation study, we utilized Gemma3-4B~\cite{gemmateam2025gemma3technicalreport} to investigate whether ECG knowledge in language models influences pretraining effectiveness. By examining its generated outputs, we found that Gemma3-4B possesses limited ECG knowledge related to ECGs.

\subsection{Dataset}
The dataset consisted of 33,732 ECG data from our hospital. The ECG data consisted of 12-lead recordings sampled at 500 Hz over a duration of 10 seconds. We split the data such that the label distribution is uniform across the training, validation, and test sets. We ensured that the same patient did not appear across different splits, as this could lead to data leakage.
The training text was formatted as: “This ECG shows \{finding\_1\}, \{finding\_2\}, …, \{finding\_n\}.”







\section{Experiments}
We conducted a series of experiments for comparison. In the first experiment, we followed the standard SigLIP training process. In the second experiment, we modified the loss function of the standard SigLIP to account for the multi-label nature specific to ECG data. While SigLIP trains by treating diagonal pairs as the correct labels, ECG datasets with a limited number of diagnostic categories may contain patients with the same ECG findings within the same batch, which can lead to label conflicts. To address this issue, we modified the loss function. The modified loss was designed to treat patients with the same condition as similar pairs, and the loss calculation was adjusted accordingly to account for this similarity. We used the Jaccard Score as a metric for this similarity. Further details are provided in the Appendix~\ref{appendix:loss}.

For all two experiments, training was conducted using the Adam optimizer with a learning rate of \(1 \times 10^{-3}\). The models were trained for 250~epochs, with a warm-up phase of 5{,}000~steps.

The results are presented in Table~\ref{result:exp1,result:exp2}. 
Evaluation was performed using the multi-label metrics: Hamming Loss, Precision~(Micro), Recall~(Micro), F1~Score~(Micro), and Jaccard~Index.

\begin{table}[htbp]
\centering
\caption{Results of the standard SigLIP and SigLIP with the modified loss}
\label{result:exp1,result:exp2}
\begin{tabular}{lcc}
\hline
\textbf{Metric} & \textbf{Standard} & \textbf{Modified loss} \\
\hline
Hamming Loss & 0.0665\,$\downarrow$ & 0.0451\,$\downarrow$ \\
Precision (Micro) & 0.5067\,$\uparrow$ & 0.3147\,$\uparrow$ \\
Recall (Micro) & 0.0365\,$\uparrow$ & 0.3020\,$\uparrow$ \\
F1 Score (Micro) & 0.0681\,$\uparrow$ & 0.3082\,$\uparrow$ \\
Jaccard Index & 0.0373\,$\uparrow$ & 0.0858\,$\uparrow$ \\
\hline
\end{tabular}
\end{table}

From Table~\ref{result:exp1,result:exp2}, it can be observed that the Modified Loss exhibits superior performance in multi-label ECG classification, as indicated by metrics such as F1~Score~(Micro), Jaccard~Index, and Hamming~Loss.

In the third experiment, we trained SigLIP using a language model without ECG-related knowledge to investigate how the presence or absence of domain knowledge in the language model affects pretraining performance. In all subsequent experiments, we employ our Jaccard-based sigmoid loss function instead of the original sigmoid loss of SigLIP.

\begin{table}[htbp]
\centering
\caption{Results of SigLIP with the modified loss, and Gemma3-4b}
\label{result:exp2,result:exp3}
\begin{tabular}{lcc}
\hline
\textbf{Metric} &  \textbf{Modified loss~(Qwen3-8B)} & \textbf{Gemma3-4b} \\
\hline
Hamming Loss & 0.0451\,$\downarrow$ & 0.0539\,$\downarrow$ \\
Precision (Micro) & 0.3147\,$\uparrow$ & 0.2451\,$\uparrow$ \\
Recall (Micro) & 0.3020\,$\uparrow$ & 0.2970\,$\uparrow$ \\
F1 Score (Micro) & 0.3082\,$\uparrow$ & 0.2686\,$\uparrow$ \\
Jaccard Index & 0.0858\,$\uparrow$ & 0.0736\,$\uparrow$ \\
\hline
\end{tabular}
\end{table}

From the results in Table~\ref{result:exp2,result:exp3}, it can be seen that the medical knowledge of the language model affects the overall performance of multi-label classification.

Through the experiments conducted thus far, we have demonstrated that employing the Modified Sigmoid Loss, which is tailored for multi-label classification, together with a language model incorporating medical knowledge, leads to performance improvements. However, the overall F1~Score~(Micro) remains low at 0.3082, which is insufficient for practical applications. 

To further enhance the F1~Score~(Micro), we conducted several performance improvement experiments. The first approach involved increasing the dimensionality of the embedding vector, which represents the final similarity, from 128 to 256. 
The reason for increasing the embedding dimensionality is that 128 dimensions may be insufficient to adequately capture the representations of ECG signals. We also experimented with 512 dimensions, but no further performance improvement was observed; therefore, those results are omitted.
The second approach aimed to address the issue of data drift by randomly cropping ECG waveforms. Since real ECG signals may vary in both start and end times, this variability can degrade performance. By applying random cropping, we mitigate this issue. 

In addition, to ensure that the effect of random cropping is properly reflected in the model, we set the warmup steps to 20,000, following the original SigLIP paper, and increased the number of training epochs to 600.

\begin{table}[htbp]
\centering
\caption{Performance comparison of baseline and proposed enhancements}
\label{result:performance_improvements}
\begin{tabular}{lcccc}
\hline
\textbf{Metric} 
& \textbf{Baseline} 
& \makecell{\textbf{Embedding}\\\textbf{dim 256}} 
& \makecell{\textbf{Embedding dim 256}\\\textbf{+ random crop}\\\textbf{(250~epoch, 5k warmup)}} 
& \makecell{\textbf{Embedding dim 256}\\\textbf{+ random crop}\\\textbf{(600~epoch, 20k warmup)}} \\
\hline
Hamming Loss & 0.0451\,$\downarrow$ & 0.0769\,$\downarrow$ & 0.0856\,$\downarrow$ & 0.0680\,$\downarrow$ \\
Precision (Micro) & 0.3147\,$\uparrow$ & 0.4097\,$\uparrow$ & 0.3824\,$\uparrow$ & 0.4898\,$\uparrow$ \\
Recall (Micro) & 0.3020\,$\uparrow$ & 0.3521\,$\uparrow$ & 0.4636\,$\uparrow$ & 0.5165\,$\uparrow$ \\
F1 Score (Micro) & 0.3082\,$\uparrow$ & 0.3788\,$\uparrow$ & 0.4191\,$\uparrow$ & 0.5028\,$\uparrow$ \\
Jaccard Index & 0.0858\,$\uparrow$ & 0.2218\,$\uparrow$ & 0.2827\,$\uparrow$ & 0.3495\,$\uparrow$ \\
\hline
\end{tabular}
\end{table}

The results are presented in Table~\ref{result:performance_improvements}. As a result, the final F1~Score~(Micro) increased to 0.5028. Although the type and amount of data differ, this result achieves an F1-score comparable to that reported in the prior CLIP-based study~\cite{li2023frozenlanguagemodelhelps}.
From these results, it can be seen that increasing the embedding dimensionality to enhance ECG representation and applying random cropping to address data drift both contribute to improved multi-label prediction performance when training ECGs with SigLIP.

We will now examine the classification performance of the final model for each individual label. The Accuracy, Precision, Recall, and F1~Score for each label are presented in Appendix Table~\ref{result:per_label_performance}. 

From this table, it can be seen that some labels are easier to train with SigLIP-based contrastive learning on ECGs, while others are more difficult. For example, findings such as ventricular premature contractions and myocardial infarction have low F1~scores, indicating that they are difficult to predict from ECGs. Additionally, conditions observable via echocardiography, such as left atrial enlargement and left ventricular hypertrophy, have relatively low accuracy, showing that it is challenging to predict them without any misclassification. In contrast, labels such as atrial fibrillation, ST-T abnormalities, and right and left bundle branch blocks are easier to predict from ECGs. Additionally, for lowEF, which is a condition observable via echocardiography, the model achieves a high accuracy of 0.9138 and an F1 Score of 0.5152. Furthermore, as shown in Appendix~\ref{appendix:roc_curves}, lowEF achieved a high AUC of 0.887, confirming its strong average predictive performance. This indicates that SigLIP is capable of predicting certain conditions, such as lowEF, which are typically identified from echocardiography, directly from ECG data.

We investigated whether performance degradation occurs when using ECG data obtained from a different hospital. The results are presented in Appendix~\ref{appendix:eval_different_institution}. Overall, the F1 score decreased only slightly to 0.4841, a reduction of approximately 0.02, indicating minimal decline in the model’s inference performance. Predictions for conditions such as lowEF also maintained an AUC of 0.888. These results suggest that our training approach is capable of preserving performance even on data from a different medical institution.

Furthermore, the experiments with the ResNet1D multi-label model are presented in Appendix~\ref{appendix:comparison_multilabel_resnet}. In these experiments as well, our model demonstrated superior performance.











\section{Conclusion}
In this study, we enhanced the performance of multi-label electrocardiogram (ECG) classification by employing a SigLIP-based ECG encoder trained on real-world clinical data and a modified loss function incorporating the Jaccard similarity. By increasing the embedding dimension and applying random cropping, the F1 score improved to 0.50, revealing which findings are relatively easy or difficult to predict. These results contribute to establishing a foundation for multimodal medical AI utilizing ECG data.

\bibliographystyle{plainnat}   
\bibliography{references}

\newpage
\section{Appendix}
\subsection{Labels}\label{appendix:labels}
In this study, the labels used for training is selected under the guidance of the cardiologists. These labels are listed in Table~\ref{tab:ecg_findings}. Note that the ground truth for lowEF, left ventricular hypertrophy, and left atrial enlargement was obtained from echocardiography data not than from ECG.
\begin{table}[htpb]
  \caption{ECG findings used in this study}
  \label{tab:ecg_findings}
  \centering
  \begin{tabular}{l}
    \toprule
    ECG Findings \\
    \midrule
    Left ventricular hypertrophy \\
    Left atrial enlargement \\
    Low ejection fraction (lowEF) \\
    Normal range (Normal) \\
    Prolonged QT interval \\
    Tall T wave \\
    Left axis deviation \\
    Artificial pacemaker rhythm \\
    Intraventricular conduction delay \\
    Complete right bundle branch block \\
    Complete left bundle branch block \\
    Flat T wave \\
    Inverted T wave \\
    ST-T abnormality \\
    Poor R wave progression \\
    Abnormal Q wave \\
    Anterior wall myocardial infarction \\
    Lateral wall myocardial infarction \\
    Inferior wall myocardial infarction \\
    Anterior septal myocardial infarction \\
    Ventricular premature contraction \\
    Frequent ventricular premature contraction \\
    Ventricular bigeminy \\
    Ventricular tachycardia \\
    Couplet of ventricular premature contractions \\
    Atrial fibrillation \\
    \bottomrule
  \end{tabular}
\end{table}
\subsection{Modified sigmoid loss}\label{appendix:loss}

We improved the original loss (Listing~\ref{alg:sigmoid_loss}) to enhance multi-label prediction performance. 

\begin{lstlisting}[style=mystyle, caption={Original Sigmoid loss pseudo-implementation.}, label={alg:sigmoid_loss}][htpb]
# img_emb     : image model embedding [n, dim]
# txt_emb     : text model embedding [n, dim]
# t_prime, b  : learnable temperature and bias
# n           : mini-batch size

t = exp(t_prime)
zimg = l2_normalize(img_emb)
ztxt = l2_normalize(txt_emb)
logits = dot(zimg, ztxt.T) * t + b
labels = 2 * eye(n) - ones(n)  # -1 with diagonal 1
l = -sum(log_sigmoid(labels * logits)) / n
\end{lstlisting}

Specifically, we modified the \textit{eye} component in Listing~\ref{alg:sigmoid_loss}. 
The original \textit{eye} is defined as a diagonal matrix 
\begin{equation}
\label{eq:eye}
\mathrm{eye} = \{ E \in \{0,1\}^{n \times n} \mid E_{ii} = 1,\; E_{ij} = 0 \; (i \neq j) \},
\end{equation}
that is, a matrix whose diagonal entries are one and off-diagonal entries are zero. 
The entries of one correspond to positive labels, whereas the zeros represent negative labels. 
This implies that the $i$-th ECG finding is considered positive only for the $i$-th label.

However, it can easily occur that the patients with the same diseases are included in the same batch. 
We then modified the \textit{eye} in Eq.~\ref{eq:eye} based on the similarity of ECG findings among patients within a batch.

We employed the Jaccard similarity to represent the similarity of these ECG findings.
The modified \textit{eye} is defined as in Eq.~\ref{eq:jaccard_similarity}, 
where the set of ECG findings for the $i$-th data is denoted by $A_i$ and 
that for the $j$-th data is denoted by $A_j$.

\begin{equation}
\label{eq:jaccard_similarity}
\mathrm{Jaccard}(A_i, A_j) = \frac{|A_i \cap A_j|}{|A_i \cup A_j|}, \quad 
\mathrm{eye}_{ij} = \mathrm{Jaccard}(A_i, A_j), \quad 
\forall i,j \in \{1, \dots, n\},
\end{equation}

The Jaccard similarity satisfies $0 \le \mathrm{Jaccard}(A_i, A_j) \le 1$, $\mathrm{Jaccard}(A_i, A_j) = \mathrm{Jaccard}(A_j, A_i)$, 
and $\mathrm{Jaccard}(A_i, A_j) = 1$ when $i = j$. 
Here, a value of $\mathrm{Jaccard}(A_i, A_j)$ closer to 1 indicates that the patients have more similar diseases. 
Using the modified \textit{eye} defined in Eq.~\ref{eq:jaccard_similarity}, 
we conducted the experiments in this study.

\subsection{Appendix: Individual Label Metrics~\ref{result:per_label_performance}}\label{appendix:label_metrics}
\begin{table}[H]
\centering
\caption{Classification performance for each label of the final model}
\label{result:per_label_performance}
\begin{tabular}{lcccc}
\hline
\textbf{Label} & \textbf{Accuracy} & \textbf{Precision} & \textbf{Recall} & \textbf{F1-Score} \\
\hline
lowEF & 0.9138 & 0.5038 & 0.5271 & 0.5152 \\
Normal & 0.9091 & 0.8054 & 0.5526 & 0.6555 \\
Prolonged QT & 0.9368 & 0.5161 & 0.2753 & 0.3590 \\
Tall T wave & 0.9842 & 0.1471 & 0.1515 & 0.1493 \\
Left axis deviation & 0.9296 & 0.3872 & 0.5884 & 0.4670 \\
Left atrial enlargement & 0.7949 & 0.4000 & 0.3336 & 0.3638 \\
Left ventricular hypertrophy & 0.7404 & 0.5932 & 0.4410 & 0.5059 \\
Artificial pacemaker rhythm & 0.9804 & 0.6564 & 0.6995 & 0.6773 \\
Intraventricular conduction delay & 0.9578 & 0.1085 & 0.1655 & 0.1311 \\
Complete right bundle branch block & 0.9674 & 0.8351 & 0.7607 & 0.7962 \\
Complete left bundle branch block & 0.9737 & 0.4138 & 0.8571 & 0.5581 \\
Flat T wave & 0.8808 & 0.5251 & 0.5849 & 0.5534 \\
Inverted T wave & 0.9355 & 0.5065 & 0.4140 & 0.4556 \\
ST-T abnormality & 0.9122 & 0.8242 & 0.6239 & 0.7102 \\
Poor R wave progression & 0.9339 & 0.4179 & 0.6062 & 0.4947 \\
Abnormal Q wave & 0.9553 & 0.0234 & 0.0420 & 0.0300 \\
Anterior wall myocardial infarction & 0.9314 & 0.0422 & 0.3226 & 0.0746 \\
Lateral wall myocardial infarction & 0.9423 & 0.0382 & 0.2778 & 0.0671 \\
Inferior wall myocardial infarction & 0.9380 & 0.1887 & 0.4348 & 0.2632 \\
Anterior septal myocardial infarction & 0.9426 & 0.1771 & 0.4551 & 0.2549 \\
Ventricular premature contraction & 0.9163 & 0.3149 & 0.3242 & 0.3195 \\
Frequent ventricular premature contraction & 0.9751 & 0.4298 & 0.3190 & 0.3662 \\
Ventricular bigeminy & 0.9777 & 0.1020 & 0.3409 & 0.1571 \\
Ventricular tachycardia & 0.9665 & 0.0000 & 0.0000 & 0.0000 \\
Couplet of ventricular premature contractions & 0.9672 & 0.0314 & 0.1034 & 0.0482 \\
Atrial fibrillation & 0.9685 & 0.8971 & 0.8700 & 0.8833 \\
\hline
\end{tabular}
\end{table}

\subsection{Appendix: ROC curves}\label{appendix:roc_curves}

\includegraphics[width=0.32\textwidth]{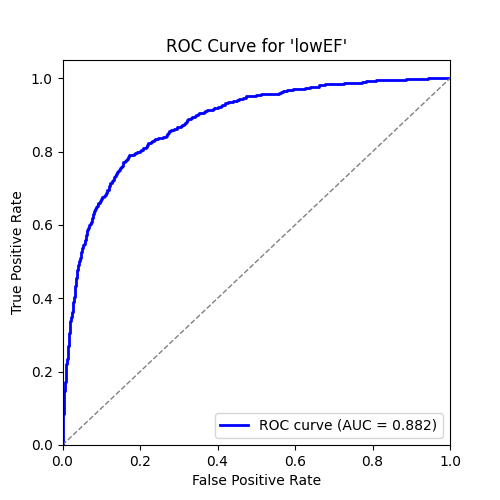}
\includegraphics[width=0.32\textwidth]{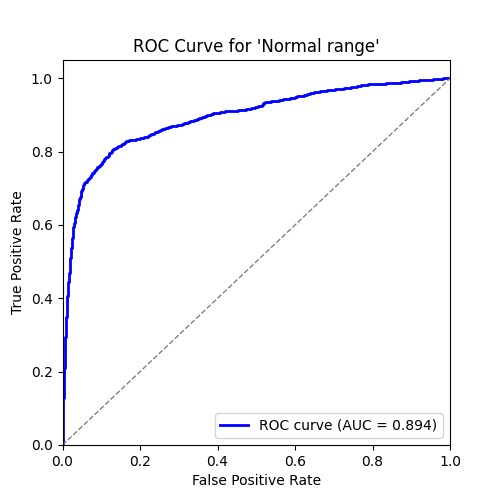}
\includegraphics[width=0.32\textwidth]{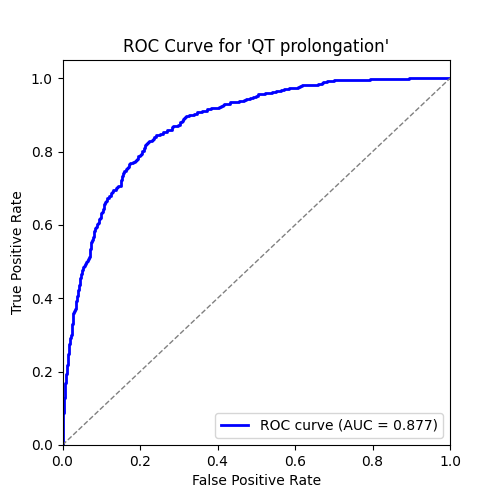}
\\
\includegraphics[width=0.32\textwidth]{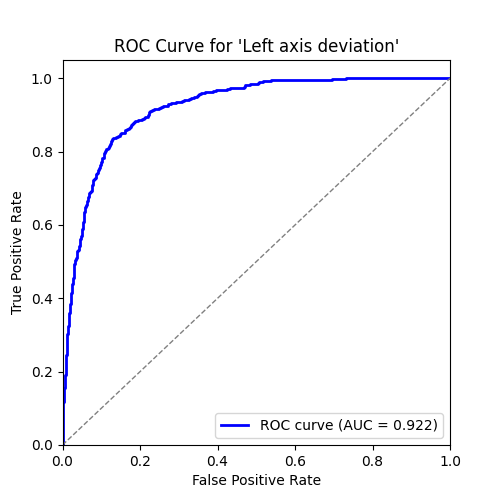}
\includegraphics[width=0.32\textwidth]{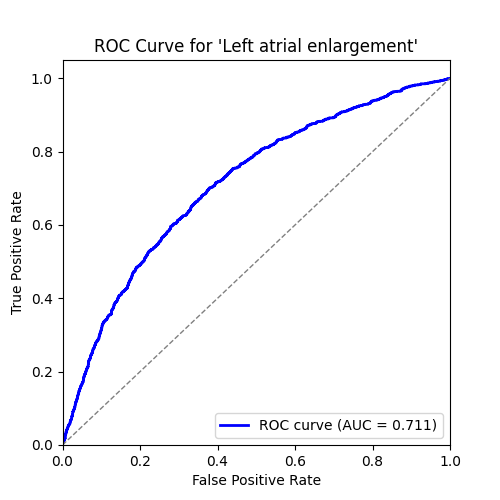}
\includegraphics[width=0.32\textwidth]{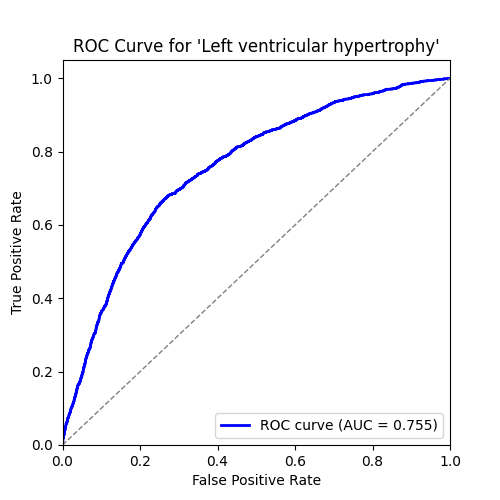}
\\
\includegraphics[width=0.32\textwidth]{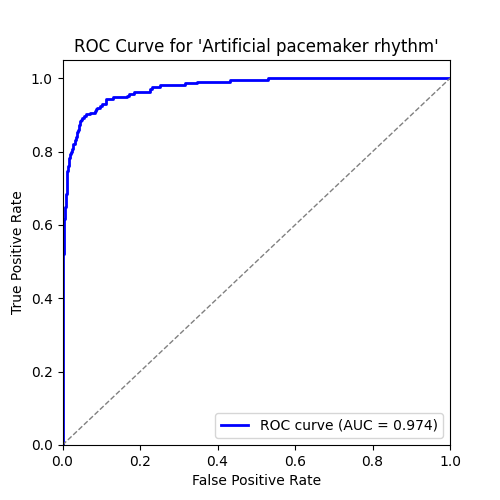}
\includegraphics[width=0.32\textwidth]{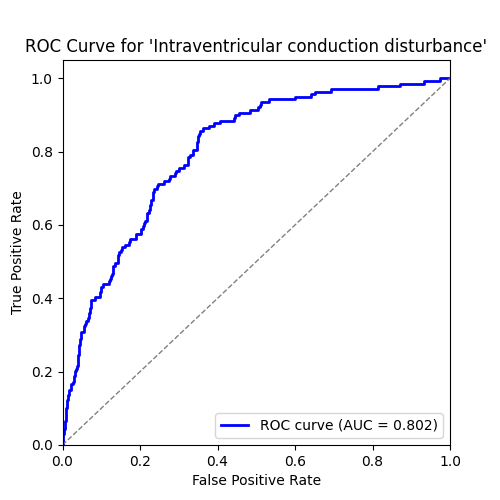}
\includegraphics[width=0.32\textwidth]{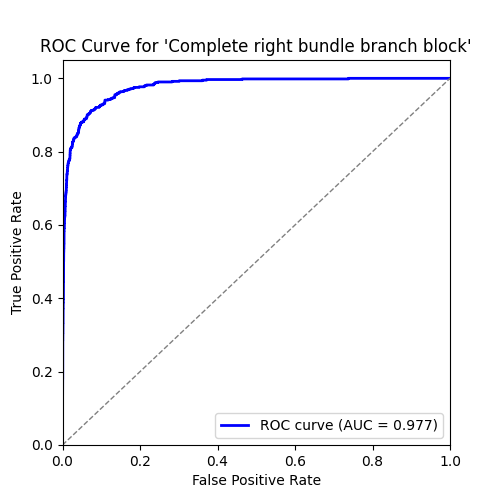}
\\
\includegraphics[width=0.32\textwidth]{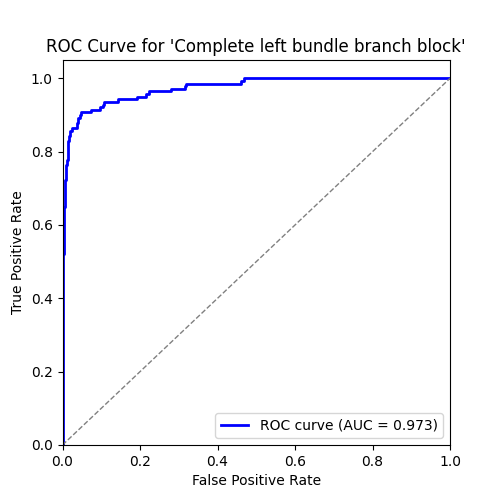}
\includegraphics[width=0.32\textwidth]{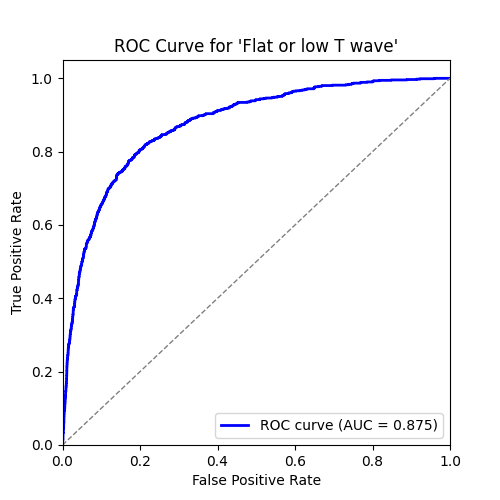}
\includegraphics[width=0.32\textwidth]{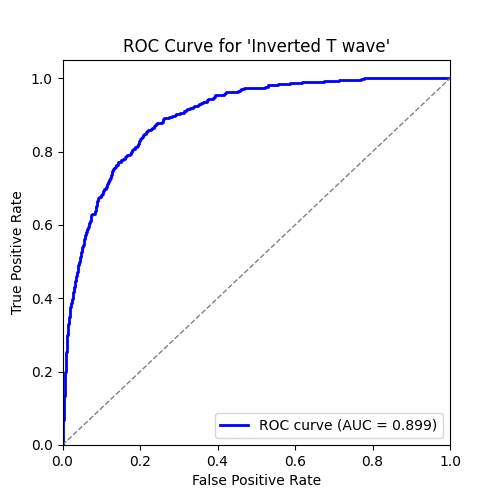}
\\
\includegraphics[width=0.32\textwidth]{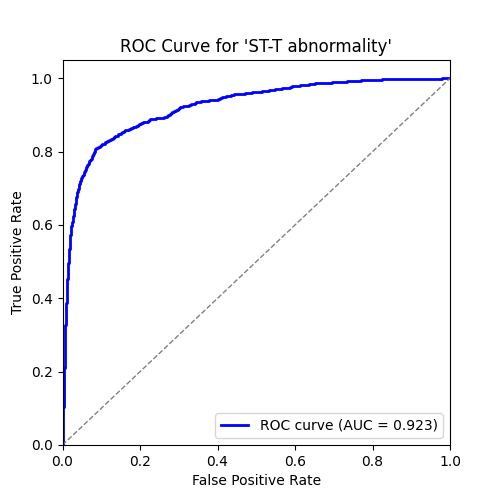}
\includegraphics[width=0.32\textwidth]{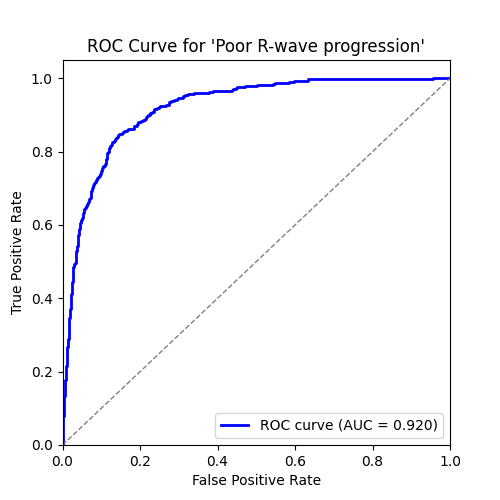}
\includegraphics[width=0.32\textwidth]{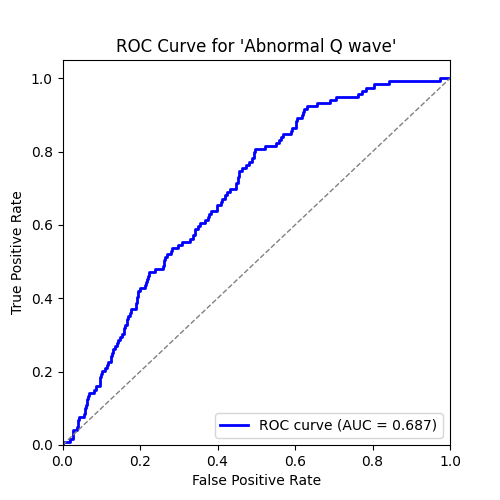}
\\
\includegraphics[width=0.32\textwidth]{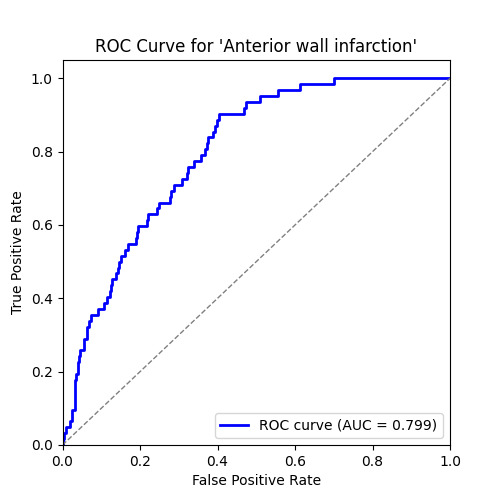}
\includegraphics[width=0.32\textwidth]{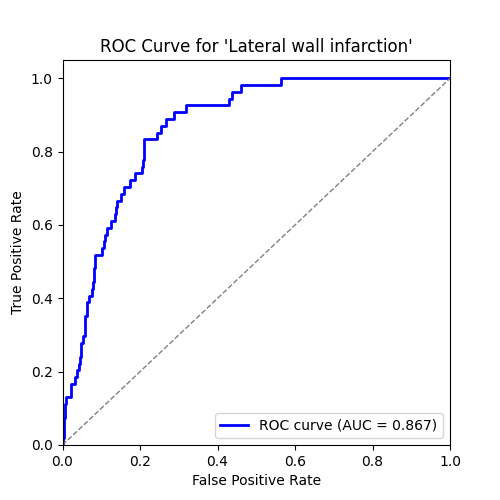}
\includegraphics[width=0.32\textwidth]{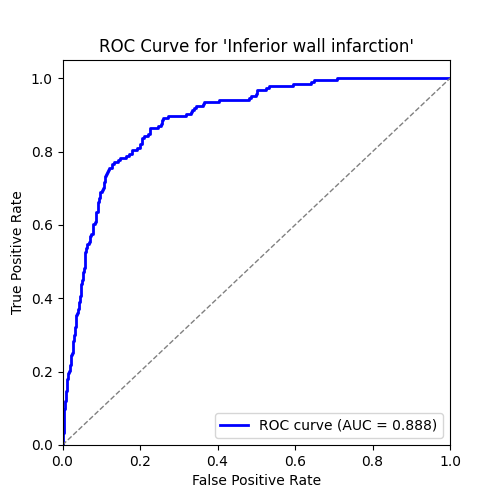}
\\
\includegraphics[width=0.32\textwidth]{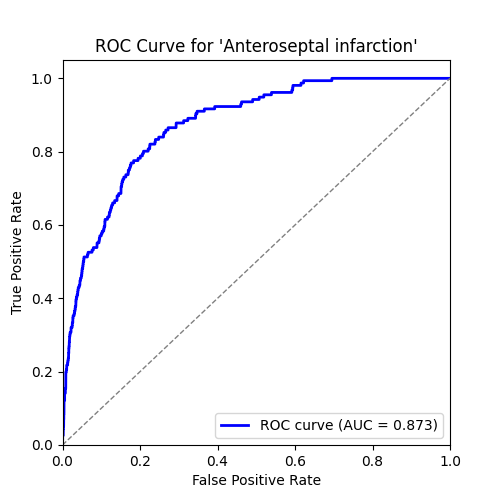}
\includegraphics[width=0.32\textwidth]{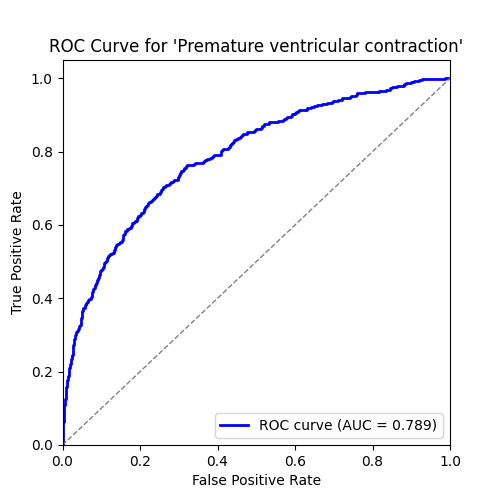}
\includegraphics[width=0.32\textwidth]{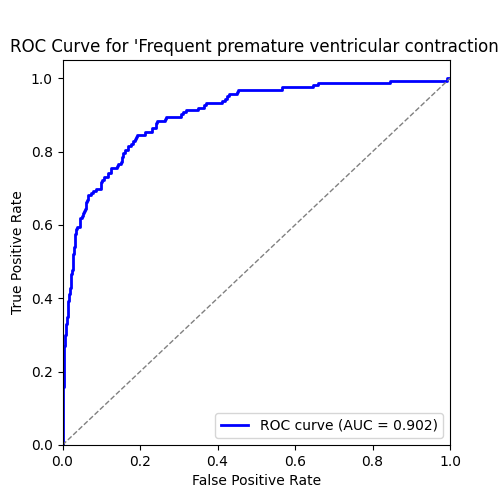}
\\
\includegraphics[width=0.32\textwidth]{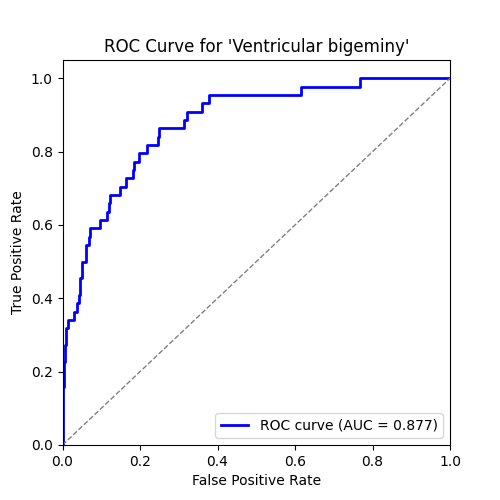}
\includegraphics[width=0.32\textwidth]{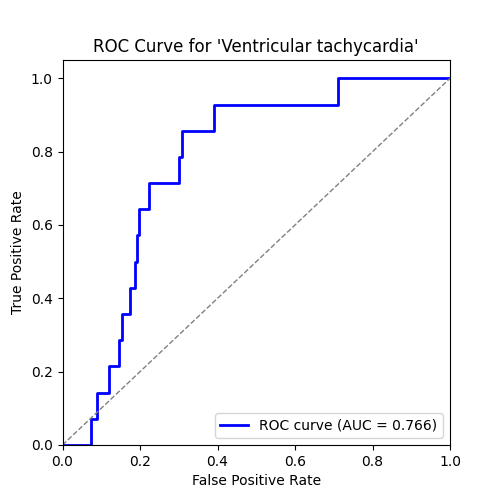}
\includegraphics[width=0.32\textwidth]{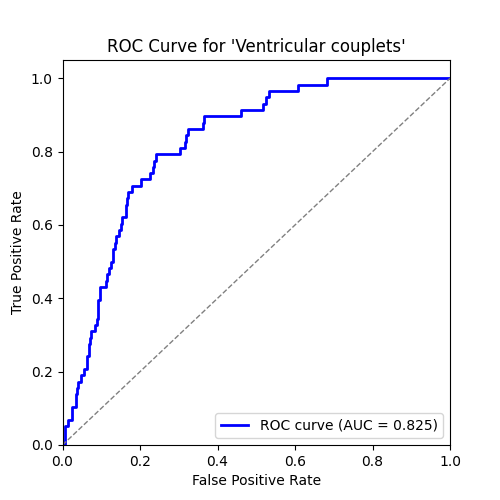}
\\
\includegraphics[width=0.32\textwidth]{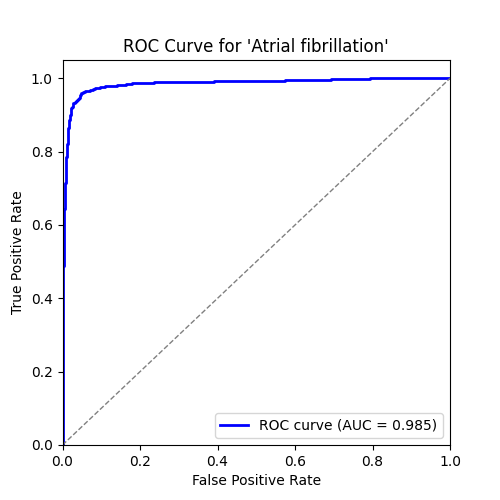}
\includegraphics[width=0.32\textwidth]{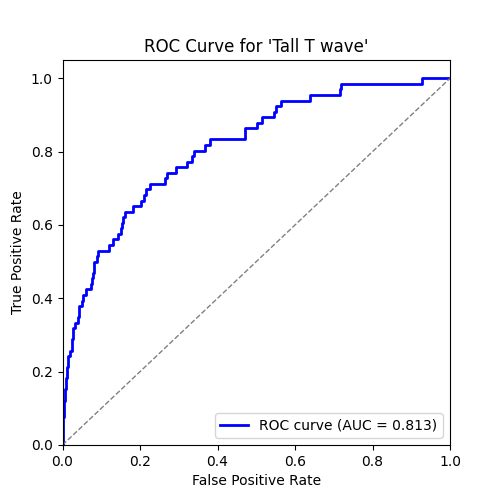}

\subsection{Evaluation on data from a different medical institution}\label{appendix:eval_different_institution}

We performed inference using data from a medical institution different from the one used for training in the paper, in order to examine the degradation in performance caused by differences in data distribution. Note that the dataset from this institution did not include any positive cases for Left Atrial enlargement or Frequent ventricular premature contractions.

\begin{table}[htbp]
\centering
\caption{Results of the different institutions}
\label{different_institutions}
\begin{tabular}{lcc}
\hline
\textbf{Metric} & \textbf{Original dataset} & \textbf{Different dataset} \\
\hline
Hamming Loss & 0.0680\,$\downarrow$ & 0.0536\,$\downarrow$ \\
Precision (Micro) & 0.4898\,$\uparrow$ & 0.4601\,$\uparrow$ \\
Recall (Micro) & 0.5165\,$\uparrow$ & 0.5107\,$\uparrow$ \\
F1 Score (Micro) & 0.5028\,$\uparrow$ & 0.4841\,$\uparrow$ \\
Jaccard Index & 0.3495\,$\uparrow$ & 0.3360\,$\uparrow$ \\
\hline
\end{tabular}
\end{table}

\begin{table}[htbp]
\centering
\caption{Classification performance for each label of different dataset}
\begin{tabular}{lcccc}
\hline
Label & Accuracy & Precision & Recall & F1-score \\
\hline
lowEF & 0.9264 & 0.5483 & 0.4504 & 0.4946 \\
Normal & 0.8793 & 0.8056 & 0.6121 & 0.6956 \\
Prolonged QT & 0.9531 & 0.3803 & 0.1421 & 0.2069 \\
Tall T wave & 0.9878 & 0.1471 & 0.1667 & 0.1563 \\
Left axis deviation & 0.9443 & 0.3804 & 0.5243 & 0.4409 \\
Left atrial enlargement & 0.9110 & 0.0000 & 0.0000 & 0.0000 \\
Left ventricular hypertrophy & 0.7525 & 0.4535 & 0.3595 & 0.4011 \\
Artificial pacemaker rhythm & 0.9660 & 0.4909 & 0.3649 & 0.4186 \\
Intraventricular conduction delay & 0.9724 & 0.0833 & 0.1905 & 0.1159 \\
Complete right bundle branch block & 0.9649 & 0.8281 & 0.6901 & 0.7528 \\
Complete left bundle branch block & 0.9812 & 0.3333 & 0.8444 & 0.4780 \\
Flat T wave & 0.8922 & 0.5204 & 0.5141 & 0.5172 \\
Inverted T wave & 0.9420 & 0.4643 & 0.4333 & 0.4483 \\
ST-T abnormality & 0.9257 & 0.7807 & 0.5848 & 0.6687 \\
Poor R wave progression & 0.9527 & 0.4007 & 0.6859 & 0.5059 \\
Abnormal Q wave & 0.9740 & 0.0172 & 0.0169 & 0.0171 \\
Anterior wall myocardial infarction & 0.9570 & 0.0407 & 0.2188 & 0.0686 \\
Lateral wall myocardial infarction & 0.9581 & 0.1043 & 0.3036 & 0.1553 \\
Inferior wall myocardial infarction & 0.9570 & 0.1477 & 0.3939 & 0.2149 \\
Anterior septal myocardial infarction & 0.9663 & 0.1489 & 0.4200 & 0.2199 \\
Ventricular premature contraction & 0.9567 & 0.4213 & 0.4601 & 0.4399 \\
Frequent ventricular premature contraction & 0.9798 & 0.0000 & 0.0000 & 0.0000 \\
Ventricular bigeminy & 0.9835 & 0.0909 & 0.3158 & 0.1412 \\
Ventricular tachycardia & 0.9703 & 0.0000 & 0.0000 & 0.0000 \\
Coupled ventricular premature contraction & 0.9740 & 0.0364 & 0.3077 & 0.0650 \\
Atrial fibrillation & 0.9783 & 0.8721 & 0.8766 & 0.8743 \\
\hline
\end{tabular}
\end{table}

\subsection{Appendix: ROC curves of different data}\label{appendix:roc_curves_different_data}

\includegraphics[width=0.32\textwidth]{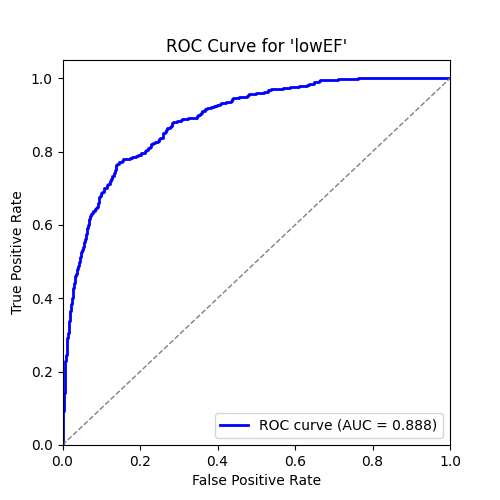}
\includegraphics[width=0.32\textwidth]{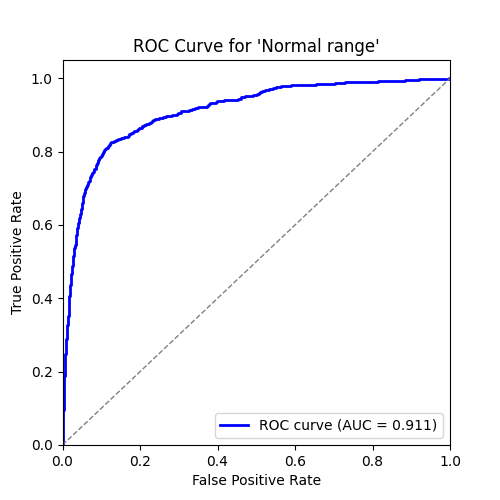}
\includegraphics[width=0.32\textwidth]{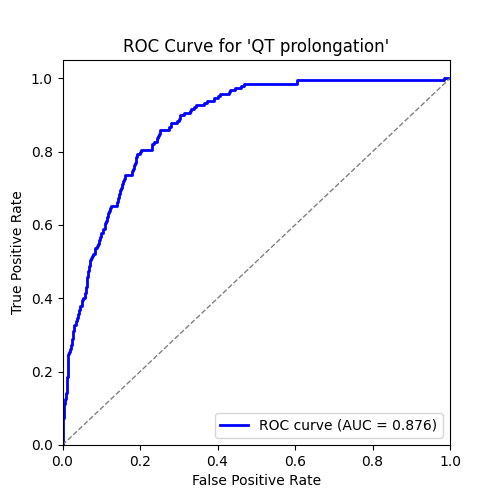}
\\
\includegraphics[width=0.32\textwidth]{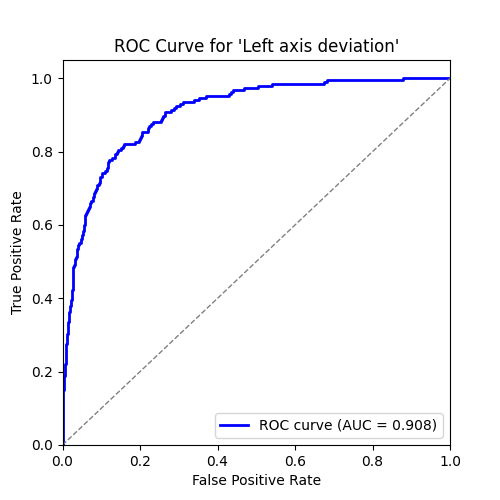}
\includegraphics[width=0.32\textwidth]{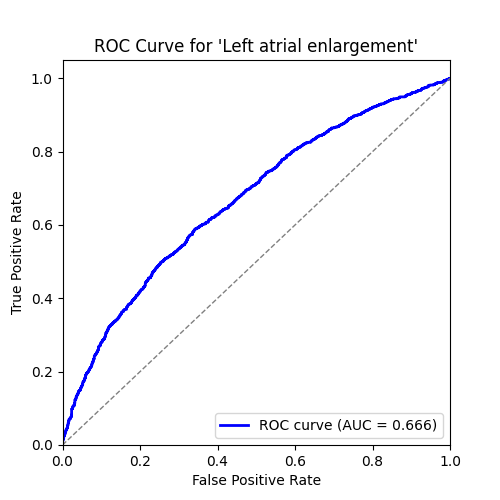}
\includegraphics[width=0.32\textwidth]{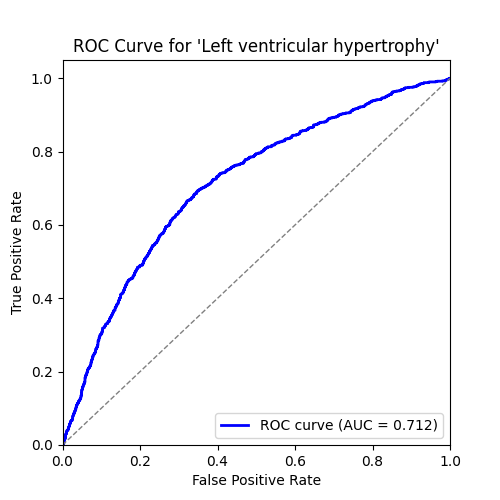}
\\
\includegraphics[width=0.32\textwidth]{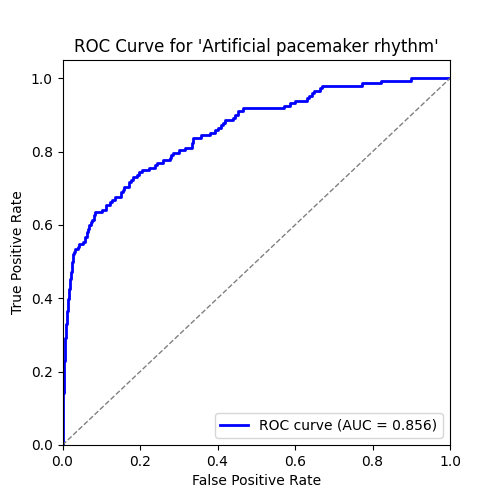}
\includegraphics[width=0.32\textwidth]{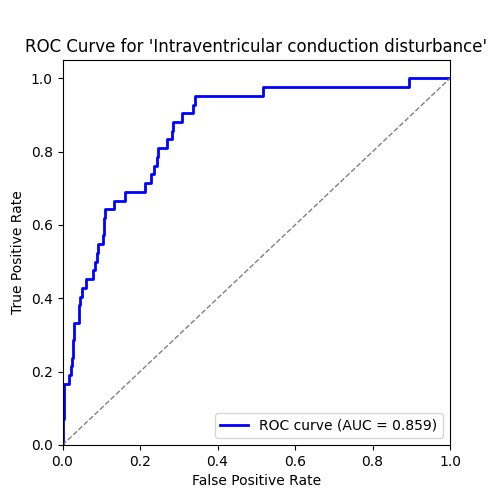}
\includegraphics[width=0.32\textwidth]{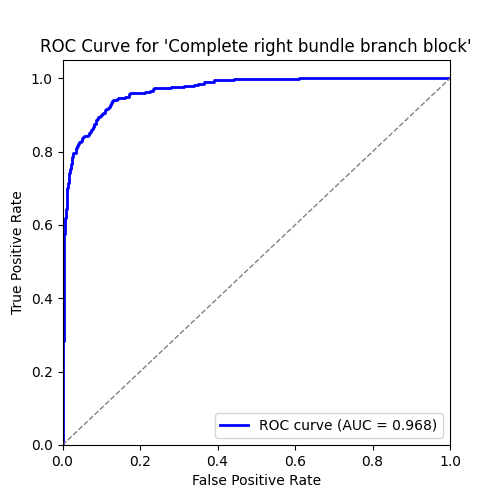}
\\
\includegraphics[width=0.32\textwidth]{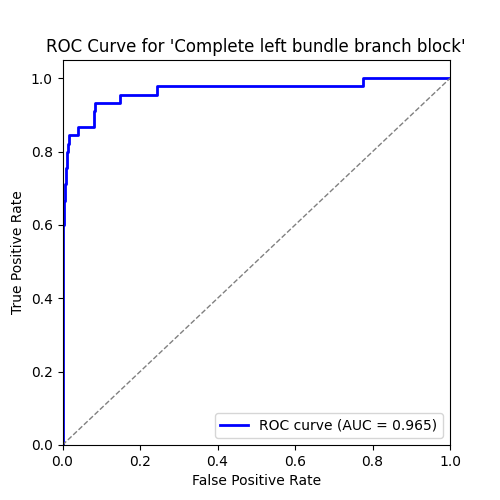}
\includegraphics[width=0.32\textwidth]{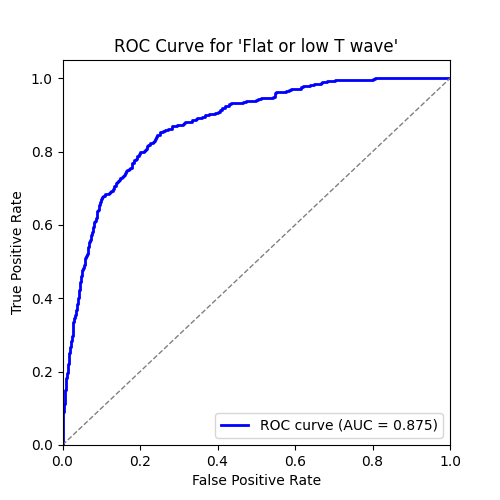}
\includegraphics[width=0.32\textwidth]{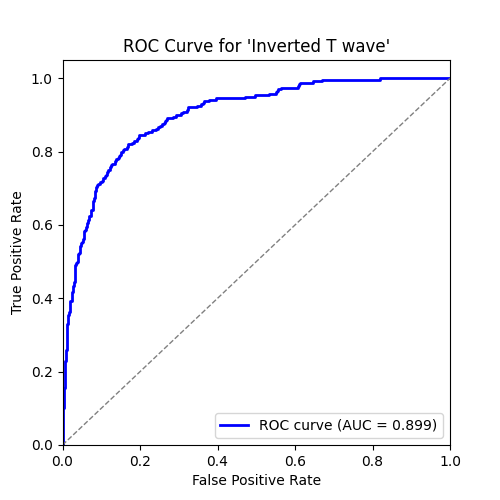}
\\
\includegraphics[width=0.32\textwidth]{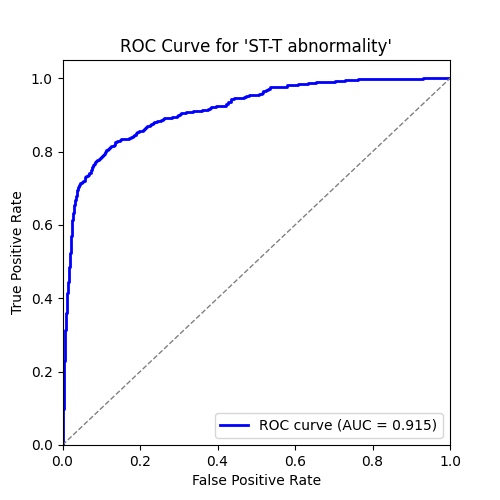}
\includegraphics[width=0.32\textwidth]{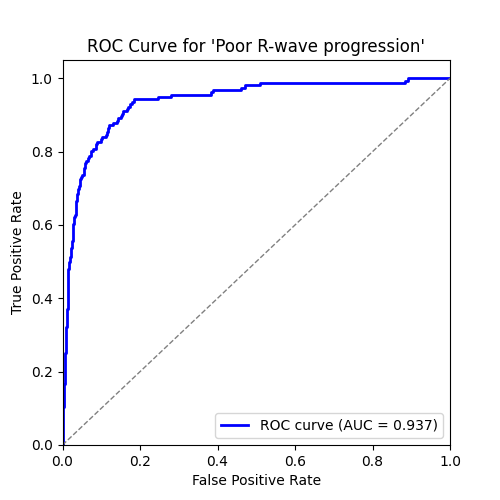}
\includegraphics[width=0.32\textwidth]{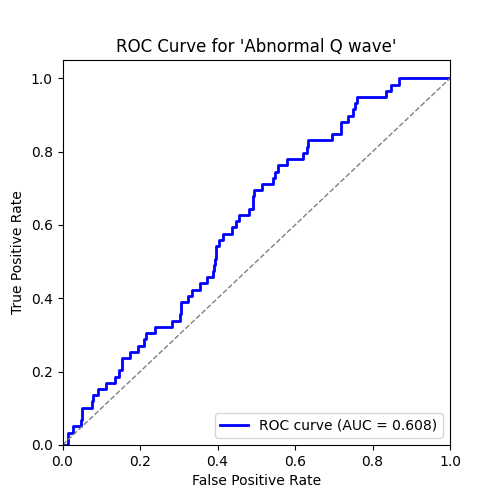}
\\
\includegraphics[width=0.32\textwidth]{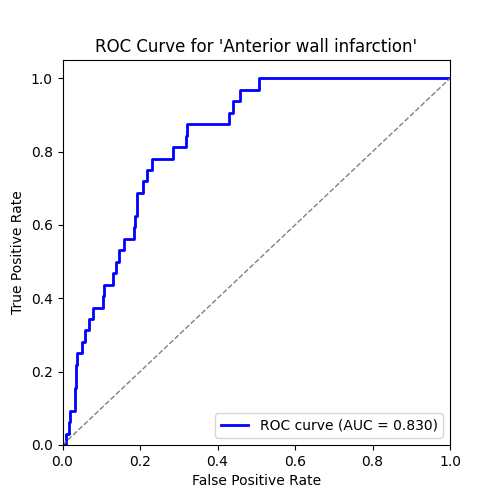}
\includegraphics[width=0.32\textwidth]{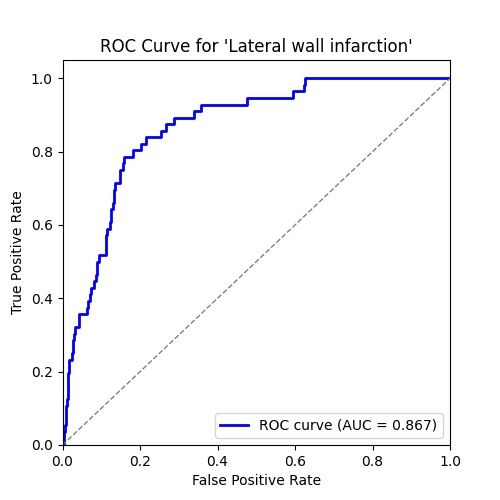}
\includegraphics[width=0.32\textwidth]{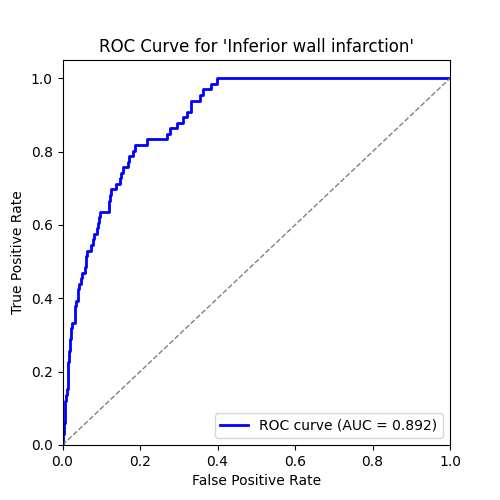}
\\
\includegraphics[width=0.32\textwidth]{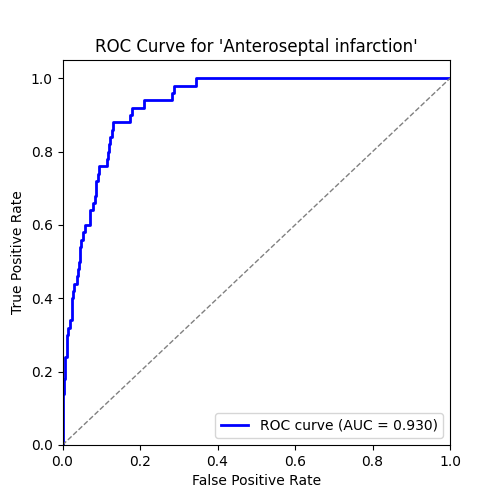}
\includegraphics[width=0.32\textwidth]{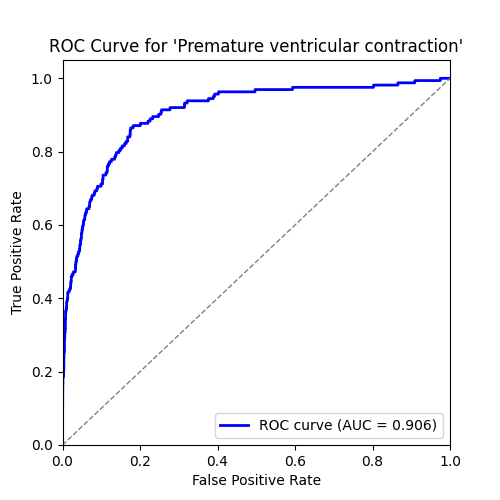}
\includegraphics[width=0.32\textwidth]{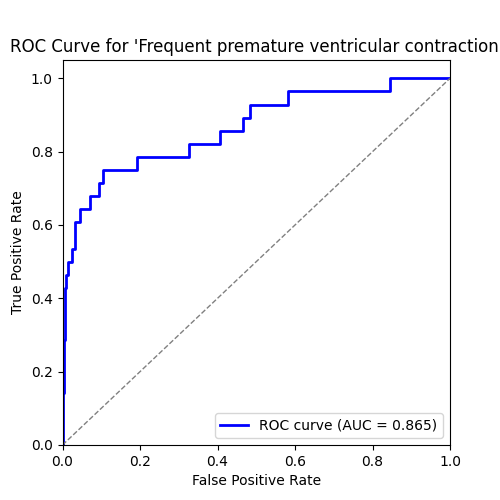} 
\\
\includegraphics[width=0.32\textwidth]{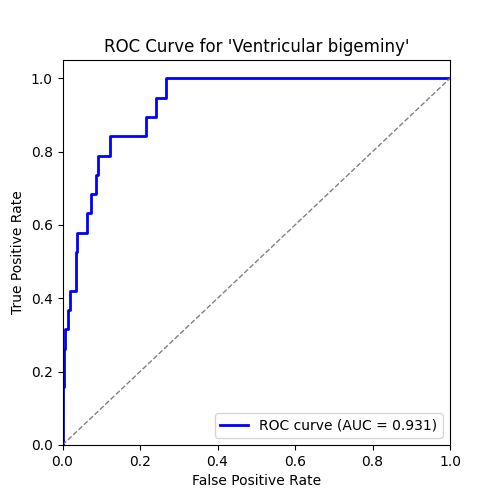}
\includegraphics[width=0.32\textwidth]{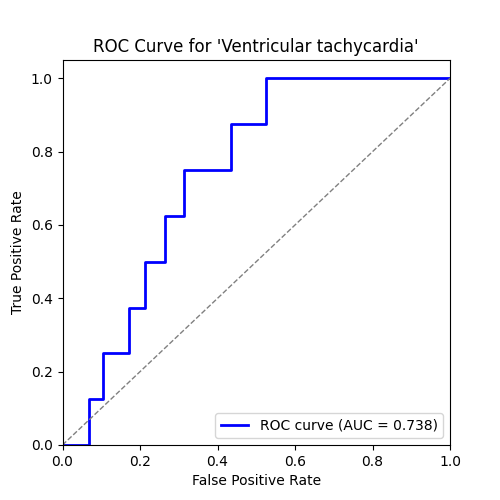}
\includegraphics[width=0.32\textwidth]{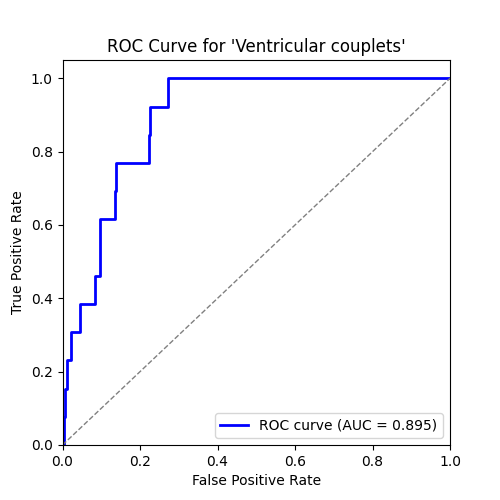}
\\
\includegraphics[width=0.32\textwidth]{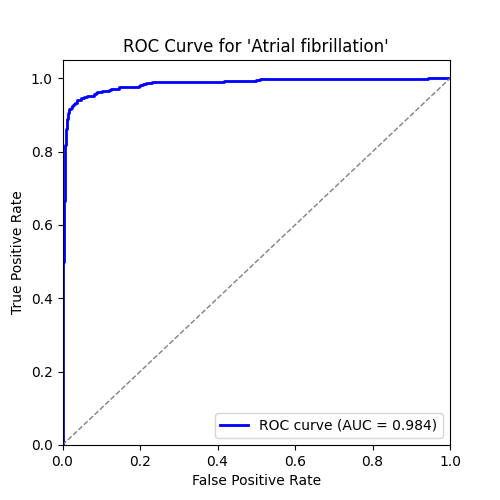}
\includegraphics[width=0.32\textwidth]{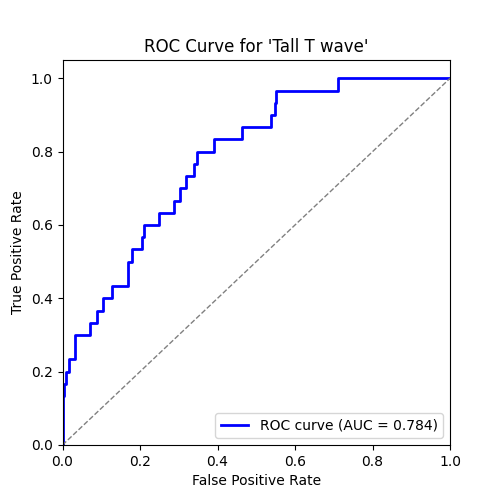}

\subsection{Ablation Study: Comparison with a ResNet-Based Multilabel Classification Model}\label{appendix:comparison_multilabel_resnet}
As an ablation study, we used a ResNet-1D model for multilabel prediction. The model was trained for 600 epochs with a learning rate of 1e-4 and a batch size of 32. As described, we applied data augmentation to the training set using random cropping. Since training can become unstable when only a small number of samples are available for certain labels, we weighted the loss according to the label distribution.

\begin{table}[htbp]
\centering
\caption{Performance comparison of baseline and a ResNet-based multi-label model}
\label{result:ablation_study}
\begin{tabular}{lcccc}
\hline
\textbf{Metric} 
& \makecell{\textbf{SigLIP Embedding dim 256}\\\textbf{+ random crop}\\\textbf{(600~epoch, 20k warmup)}}
& \makecell{\textbf{ResNet-based multi-label model}\\\textbf{ + random crop}} \\
\hline
Hamming Loss & 0.0680\,$\downarrow$ & 0.1854\,$\downarrow$ \\
Precision (Micro) & 0.4898\,$\uparrow$ & 0.2444\,$\uparrow$ \\
Recall (Micro) & 0.5165\,$\uparrow$ & 0.8526\,$\uparrow$ \\
F1 Score (Micro) & 0.5028\,$\uparrow$ & 0.3799\,$\uparrow$ \\
Jaccard Index & 0.3495\,$\uparrow$ & 0.2641\,$\uparrow$ \\
\hline
\end{tabular}
\end{table}
As shown in Table~\ref{result:ablation_study}, the proposed SigLIP-based method achieved a higher F1 score than a ResNet-based multi-label classification. This improvement is likely because multimodal training allows the model to leverage features embedded in the text, enabling more accurate inference even with a smaller number of ECG findings.

\end{document}